\documentclass[final,3p,times]{elsarticle}

\usepackage{amssymb}
\usepackage{tcolorbox}
\usepackage{mathtools}
\usepackage{float}
\usepackage{array}
\usepackage{geometry}
\usepackage{algorithm}
\usepackage{algorithmic}
\usepackage{lipsum}
\makeatletter
\def\ps@pprintTitle{%
 \let\@oddhead\@empty
 \let\@evenhead\@empty
 \def\@oddfoot{}%
 \let\@evenfoot\@oddfoot}
\makeatother

\DeclarePairedDelimiter{\abs}{\lvert}{\rvert}

\begin{document}

\begin{frontmatter}

\title{Chaos inspired Particle Swarm Optimization with Levy Flight for Genome Sequence Assembly}

% \author{Sehej Jain}
% \ead{sehejjain@outlook.com}
% \author{Kusum Kumari Bharti}
% \ead{kusum@iiitdmj.ac.in}
% \address{PDPM Indian Institute of Information Technology Design and Manufacturing, Jabalpur\\ Madhya Pradesh, India, 482005}

\author[1]{Sehej Jain\corref{cor1}}
\ead{sehejjain@outlook.com}
\author[2]{Kusum Kumari Bharti}
\ead{kusum@iiitdmj.ac.in}
\cortext[cor1]{Corresponding author}
% \fntext[fn1]{This is the first author footnote.}
% \fntext[fn2]{Another author footnote, this is a very long
% footnote and it should be a really long footnote. But this
% footnote is not yet sufficiently long enough to make two
% lines of footnote text.}
\address[1,2]{PDPM Indian Institute of Information Technology Design and Manufacturing, Jabalpur, Madhya Pradesh, India}

\begin{abstract}
With the advent of Genome Sequencing, the field of Personalized Medicine has been revolutionized. From drug testing and studying diseases and mutations to clan genomics, studying the genome is required. However, genome sequence assembly is a very complex combinatorial optimization problem of computational biology. PSO is a popular meta-heuristic swarm intelligence optimization algorithm, used to solve combinatorial optimization problems. In this paper, we propose a new variant of PSO to address this permutation-optimization problem. PSO is integrated with the Chaos and Levy Flight (A random walk algorithm) to effectively balance the exploration and exploitation capability of the algorithm. Empirical experiments are conducted to evaluate the performance of the proposed method in comparison to the other variants of the PSO proposed in the literature. The analysis is conducted on four DNA coverage datasets. The conducted analysis demonstrates that the proposed model attain a better performance with better reliability and consistency in comparison to others competitive methods in all cases.

\end{abstract}


\end{frontmatter}


\section{Introduction}
\label{intro}
Bioinformatics and computational biology have gained immense popularity in the past decade and have seen application of various techniques from the domain of Computer Science.
Genome sequencing is one of the such field.  The main aim of genome sequencing is to determine the long DNA sequences in the genome of an organism \cite{MyersJr+2016+126+132,phillippy2017new}. The complete genome sequence information is essentially required for both basic and advanced biological research. It has numerous application in different domains such as genomics, medical diagnosis, biotechnology, virology, physiology,  cloning, forensic science etc.

 As genome mapping has marked the advent of personalized medicine, it is very important in today's and future's medical system. 
 In Next Generation Sequencing, millions of small fragments of DNA are sequenced parallelly. By mapping individual reads to the reference genome, bioinformatics methods are used to put these fragments back together. Each base in the genome is sequenced several times, resulting in a high level of precision and insight into unexplained DNA variation\cite{Behjati2013}. This combinatorial optimization problem is an NP-hard problem. Thus, heuristic and meta-heuristic algorithms are very well suited for this job. 

DNA of any being consists of three primary components:
\begin{enumerate}
    \item A Nitrogenous base
    \item A Phosphate Group
    \item A Sugar molecule
\end{enumerate}
This nitrogenous base can be one of four possible bases, adenine (A), cytosine (C), guanine (G), and thymine (T).

A DNA molecule is made up of two parallel strands that are bound by complementary molecule bonds (A\&T and G\&C). Since the human genome is 3.2 billion nucleotides long, it cannot be read all at once. Next Generation Sequencing (NGS) is used to read the genome. This method consists of cloning the DNA several times over, and then splitting all of them into millions of random fragments. This is where the sequence assembly problem lies. One of the genome sequence assembly procedures is the Overlap-Layout Consensus (OLC) approach. 

\begin{enumerate}
    \item\textbf{Overlap Stage: }
  To evaluate the similarity score, a semi-global alignment algorithm is used. It consists of comparison of all possible pairs of fragment. This technique is used to determine the overlapping fragments. It is mostly done using semi global alignments like Smith-Waterman algorithm \cite{SMITH1981195}.
    \item\textbf{Layout Stage: } 
   The exact order of fragments is found based on calculated similarity scores in this step. This was proven to be a NP-hard problem \cite{pevzner2000}.
    \item\textbf{Consensus Stage: }
   This stage helps to derive the DNA sequence from the layout stage. Traditionally, majority rule is used to build the consensus. 
\end{enumerate}

The OLC model for Genome Sequence Assembly has received much merit since being proposed in 1980 \cite{Staden1980,Sanger1982}. We only dwell on the layout phase in this paper because it is at the heart of the DNA fragment assembly problem.

\begin{figure}
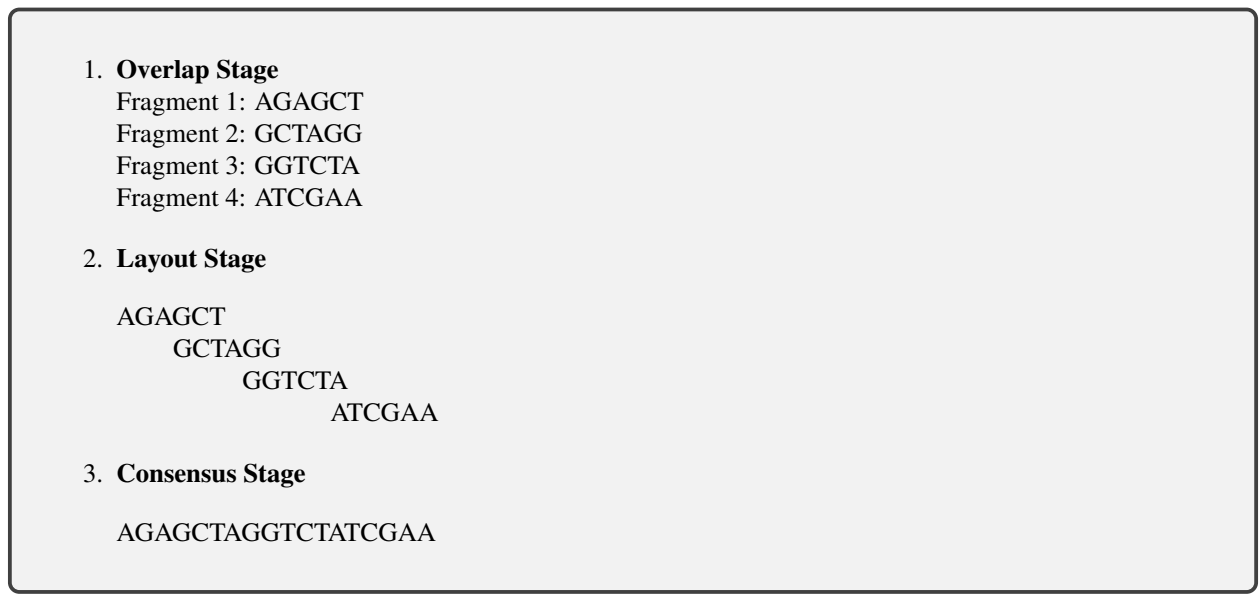

\centering
\begin{tcolorbox}
\begin{enumerate}
\vspace{1em}
    \item \textbf{Overlap Stage}
    \\Fragment 1: AGAGCT
    \\Fragment 2: GCTAGG
    \\Fragment 3: GGTCTA
    \\Fragment 4: ATCGAA
    \vspace{1em}
    \item \textbf{Layout Stage}
    \vspace{1em}
    \\AGAGCT
    \\\phantom{AGA}GCTAGG
    \\\phantom{AGAGCTA}GGTCTA
    \\\phantom{AGAGCTAGGTCT}ATCGAA
    \vspace{1em}
    \item \textbf{Consensus Stage}
    \vspace{1em}
    \\AGAGCTAGGTCTATCGAA
     \vspace{1em}
\end{enumerate}
\end{tcolorbox}
\caption{Genome Sequence Assembly Example}
\label{fig:genomeAssembly}
\end{figure}

Figure \ref{fig:genomeAssembly} shows how the OLC model for Genome Sequence Assembly works with an example.

The first work on identification of the presence of ‘nuclein’ (currently, known as DNS) has been published by Friedrich Miesche\cite{MyersJr2016} in 1871. Fast forward to 1990, when the Human Genome Project was launched, aimed to deduce the entire Human Genome. It was completed in 2003, with an accuracy of 99.99\%. Various genetic algorithms were analysed in this project for this permutation-based problem.

As genome sequencing is a combinatorial optimization problem, different nature inspired algorithms have been proposed in past few decades to solve this problem \cite{NEBRO2008,Alba2008,Blazeewicz2004,Allaoui2018}. A. J. Nebro et al \cite{NEBRO2008} had presented a grid based genetic algorithm for reconstructing and to solve the subsequence overlap problem. In 2008, Gabriel Luque and Enrique Alba proposed a hybrid Genetic Algorithm \cite{Alba2008} for Genome Sequencing. Since then various algorithms\cite{Baichoo2017} have been proposed in the literature to solve this problem. Ant Colony Optimization \cite{Meksangsouy2003,Wetcharaporn2006,Ibrahim2009}, Genetic Algorithm\cite{Fang2005,Parsons1995,Alba2008}, Artificial Bee Colony\cite{Firoz2012}, Hierarchy Clustering, Simulated Annealing, Firefly Algorithm\cite{Vidal2018}, Tabu Search\cite{Blazeewicz2004}, De Novo Assembly Algorithm\cite{Denovo2017} and Crow Search Algorithm \cite{Allaoui2018} techniques have been leveraged too.

In this paper, a Shortest Position Value rule is used to tackle Genome Sequencing which is a permutation optimization problem \cite{ShankarVerma2011}.
DNA Subsequence Assembly using Probabilistic Edge Recombination (PER) has also been proposed with Particle Swarm Optimization\cite{BenAli2020} to solve this problem.
In the past few years, algorithms such as  Cuckoo Search have been used too \cite{Indumathy2014,Karthik2018,Raja2018}. A hybrid approach using PSO and Grey Wolf Optimization was proposed by Xinming Zhang et al \cite{ZHANG2021}.
PSO has also been the subject of focus for many Genome Sequence Assembly papers\cite{Hefny2010,Huang2015,Liu2005}.

Some of the challenges that the Genome Sequence Assembly problem entails are:
\begin{enumerate}
    \item The size of data can even be in the order of Terabytes, thus requiring extremely high computing power.
    \item Errors reads in the genome from the sequencing instruments can confound the assemblers and have far reaching effects on the outcome.
    \item Identical and nearly identical reads increase both the time and space complexity of the problem.
\end{enumerate}

\subsection{Paper Organisation}
\begin{enumerate}
    \item Section \ref{previous work} describes related literature's.
    \item Section \ref{background} sets a background to this paper.
    \item Section \ref{methodology} offers the details of the proposed chaos inspired PSO with Levy Flight.
    \item Section \ref{Exp Results} describes the performance and evaluation of the proposed algorithm.
    \item Section \ref{conclusion} presents brief conclusions of this paper and suggestions for future work.
\end{enumerate}
  
\section{Previous Work}
\label{previous work}
The first Genome Assemblers began to appear in the 1980's and 1990's. This section provides a brief overview of Genome Sequencing techniques and related literature.
\subsection{Biological Process}
This section presents the Biological processes and techniques that are relevant to Genome Sequencing.
\subsubsection{First Generation Sequencing}
Allan Maxam and Walter Gilbert proposed First Generation Sequencing, commonly known as Maxam-Gilbert Sequencing, in 1970. It was based on nucleobase-specific partial chemical modification of DNA, followed by cleavage of the DNA backbone at sites adjacent to the modified nucleotides. The modern Sanger DNA chain-termination sequencing method\cite{Sanger1977}, also referred as the Sanger sequencing system, triumphed over the Maxam and Gilbert chemical degradation method\cite{Maxam1977} due to it's greater flexibility and efficiency, paired with the use of less toxic substances and lower levels of radioactivity, .

\subsubsection{Next Generation Sequencing}
Next Generation Sequencing (NGS) is a technique that utilises technologies such as SOlid, Ion-Torrent, Illumina, and Roche454 sequencing. It allows scientists to sequence thousands to tens of thousands of genomes in a single year or study the entire human genome in a single sequencing experiment. Despite the fact that it is over a decade old, it is still used to characterise extremely parallel or high-output sequencing methods that generate results at or above the genome scale.
\subsection{EST Assemblers} 
EST (Express Sequence Tab) Assemblers, \cite{Chevreux2005,Chevreux2004}, sought to assemble individual genes rather than whole genomes. This differs from the genome assembly problem in that the dataset used here is a variant of the dataset used in Genome Assembly. Post-transcriptional alteration, single-nucleotide polymorphism, alternative splicing, and trans-splicing all exacerbate EST assembly.

\subsection{De-novo Assembly}

De novo gene assemblers have no reference genome for combining small nucleotide sequences into longer ones. They are widely used in Bioinformatics to solve the Sequence Assembly Problem. De-Novo Assemblers are classified into two types:
\begin{enumerate}
    \item \textbf{Greedy-based De-novo assemblers:} They find the optima in four steps\cite{Peltola1984,HUANG199218}.
    \begin{enumerate}
        \item Calculate the pairwise distance of reads
        \item Create clusters of reads with maximum overlap
        \item Assemble the overlapping reads into larger contigs
        \item Repeat until termination criteria is not satisfied
    \end{enumerate}
    \item \textbf{Graph-method assemblers:} From the advent of the Graph based assemblers it became extremely popular with next generation sequencing (NGS).
    There are two types of Graph based assemblers: String\cite{Idury1995} and De-Bruijn\cite{Compeau2011}.
    In this process,  reads are broken into $k$ pre-specified smaller fragments. In the graph assembly, the k-mers act as nodes. The overlapping nodes are then connected by an edge. Finally, the assembler builds sequences using the De Bruijn graph.
\end{enumerate}

\subsection{Mapping Assembly}
Unlike De-novo Assemblers, Mapping Assemblers\cite{Otto2015} make use of an existing backbone sequence to build a sequence that is similar however not necessarily identical to the backbone sequence.

\subsection{TIGR Assembler}
For sequencing projects, the Institute for Genomic Research developed TIGR Assembler. The TA was developed and built using the experience gained from more than 20 sequencing projects completed at TIGR.
This broad experience gave rise to a sequence assembly with few misassemblies successfully used for whole genome shotgun sequence of prokaryotic and eukaryotic organisms, artificial chromosome sequences of eukaryotic organisms and sequence tag assembly.\cite{TIGR,Pop2004}.

\subsection{EULER path approach}
Pevzner et al\cite{Pevzner2001} in 2001 proposed an alternate greedy approach to Genome Sequencing instead of the OLC approach. This approach suceeded where OLC could not show better results, in genomic shotgun assembly.

\subsection{Parallel Assemblers}
Simpson et al. \cite{Simpson2009} created ABySS (Assembly by Short Sequences). AbySS is a parallel sequence assembly process and it targeted very long sequences. For, instance, it had compiled 3.5 billion paired-end reads from an African male genome which was made public by Illumina, Inc.

\section{Background}
\label{background}
\subsection{Particle Swarm Optimization}
\label{PSO}

Particle swarm optimization (PSO) is a well-known swarm intelligence optimization algorithm used to address combinatorial optimization problems. It optimises a problem by iteratively optimising the candidate solutions in relation to a given consistency metric. Kennedy and Eberhart\cite{Kennedy1995,Eberhart1999} have developed PSO algorithm, which simulate social behaviour bird flock or fish school.
\\This algorithm excels due to its quick computations and knowledge exchange within the algorithm, since it draws its internal conclusions from particle's social activity. The particles pass through the search space, with each particle representing one potential solution to the given problem. Each particle is then given a health or score rating based on a performance-judging feature specific to the issue at hand.
\\The movement of particles is highly influenced by information from its own experience and it's neighbours' experience. Each particle stores it best value, \textit{pBest} and feels an attraction towards the global best, \textit{gBest}. These are the cognitive and social components of propagation of a particle swarm algorithm.

The position and velocity of the \(i^{th}\) particle are represented by $n$ dimensional vectors, \(X_i = (x_{i1}, x_{i2}, ...,x_{in})^T\) and \(V_i = (v_{i1}, v_{i2},...,v_{in})^T\), respectively. The previous best position of the \(i^{th}\) particle is recorded and represented by the $n$- dimensional vector, \(P_i = (p_{i1}, p_{i2}, ...,p_{in})^T\). Mathematical, formulation of the velocity update equation is given by:
\begin{equation}
    v_{id} = v_{id} + c_1r_1(p_{id}- x_{id}) + c_2r_2(p_{gd} -  x_{gd}),
\end{equation}

Here, subscript $g$ is the index of the best particle in the swarm.\\The position is then updated in each iteration using:
\begin{equation}
x_{id} = x_{id} + v_{id}
\end{equation}

 Where $d = 1,2,\cdots\cdots, n$  denotes the dimension and $i = 1,2,\cdots\cdots, S$ denotes the particle index. $S$ is the size of the swarm, $c_1$ cognitive parameter and $c_2$ social parameters. Traditionally, $r_1$ and $r_2$ are random numbers uniform drawn in the range of $0$ and $1$.

The inclusion of an inertia weight in the PSO was first proposed in the literature \cite{Eberhart1998}.
The mathematical formulation of the Inertia Weight based velocity updation is given by \begin{equation}
    v_{id} = w*v_{id} + c_1r_1(p_{id}- x_{id}) + c_2r_2(p_{gd} -  x_{gd}),
\end{equation}

Algorithm \ref{Algo:SPSO} shows the pseudo-code of Simple PSO.
\begin{algorithm}
\label{Algo:SPSO}
\caption{Simple Particle Swarm Optimization}
\begin{algorithmic} 
% \REQUIRE $n \geq 0 \vee x \neq 0$
% \ENSURE $y = x^n$
\STATE $max\_itr \leftarrow $\textit{Number of iterations}
\STATE $population \leftarrow $\textit{Swarm size}
\FOR{$i \leftarrow (1$\textit{ to population)}}
\STATE $x_i \leftarrow initialize the positopn of x^{th} particle$
\STATE $p_{best} \leftarrow x_i$
\IF{$p_{best} > g_{best} $}
\STATE $G_{best} \leftarrow x_i$
\ENDIF
\ENDFOR
\STATE $t \leftarrow 0$
\WHILE{$t < max\_itr$}
\FOR{$i \leftarrow (1$\textit{ to population)}}
\STATE $v_{id} \leftarrow v_{id} + c_1r_1(p_{id}- x_{id}) + c_2r_2(p_{gd} -  x_{gd})$
\STATE $x_i(t) \leftarrow x_i(t-1) + v_i$
\IF{$fitness(x_i) > fitness(p_{best}i)$}
\STATE $p_{best}i \leftarrow x_i$
\IF{$fitness(p_{best}i) > fitness(g_{best})$}
\STATE $g_{best} \leftarrow p_{best}i$
\STATE $g_{best} \leftarrow x_i$
\ENDIF
\ENDIF
\ENDFOR
\STATE $t \leftarrow t + 1$
\ENDWHILE
\end{algorithmic}
\end{algorithm}

\subsection{Chaos}
\label{chaos}
Chaos operates in a non-linear manner and is associated with unpredictability and complex behavior. It is highly sensitive to the initial conditions. A very small change can produce dramatically different results. Originally, it is introduced by Lorenz in 1963\cite{Lorenz1963}, since it has been utilized by various domains in science and technology, including communications, electronic circuits, robotics and others.

Two methods for altering the inertia weight of a PSO using chaos were added by Feng et al\cite{Feng2007}. One is chaotic reducing of inertia weight, and the other is chaotic random inertia weight. The latter is being considered in this article, as suggested in \cite{Hong2016}. Other works with chaos and PSO include \cite{Hefny2010,Liu2005}.
\subsection{Levy Flight}
\begin{figure}
    \includegraphics[width=0.3\textwidth]{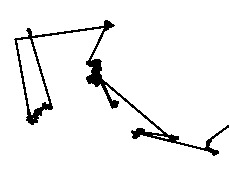}
    \caption{Simulation of Levy Flight}
    \label{fig:levy}
\end{figure}

Levy Flight\cite{Levy2008} is a random walk algorithm in which the step lengths have a heavy-tailed Levy Distribution. Named after the French Mathematician Paul Lévy, the term was coined by Benoît Mandelbrot.
Levy flight operates under the following function.
\begin{equation}
    Levy(\lambda) = \abs[\Bigg]{\frac{ \Gamma(1+\lambda) * sin(\frac{\Pi *\lambda}{2})}{\Pi\frac{(1+\lambda)}{2}* \lambda* 2^\frac{\lambda-1}{2}}} ^ \frac{1}{\lambda}
\end{equation}
The equation to find the new position is:
\begin{equation}
    X_{t+1} = X_t + S*Levy(\lambda)*\alpha
\end{equation}
where \(X_t, X_{t+1}\) are the current and new positions, \(S\) is the step size, \(\lambda\) is a constant (\(1\le\lambda\le3\)) and \(\alpha\) is a random number generated between [-1,1].
\\Figure \ref{fig:levy} shows a simulation of Levy Flight in a 2-dimensional plane.

\section{Methodology}
\label{methodology}
The Genome Sequence Assembly problem is a permutation optimization problem. Thus, standard methods of applying the PSO algorithm would not work here. One approach that allows us to use this algorithm is SPV or Shortest Position Value. We can convert our continuous position values to discrete positions using SPV\cite{Kaur2012}.

\subsection{SPV}
\label{SPV}
The position vector \(X_{id} = { x_1,x_2,x_3,......x_d }\) represents one particle, where \(i\) represents the particular particle index and \(d\) represents the dimension index. For a location optimization problem, the number of dimensions can be two or three depending on the problem.
\\In Genome Sequencing the number of dimensions is the number of fragments of a particular DNA which have to be arranged. Using the SPV (Smallest Position Value) rule\cite{Tasgetiren2009}, we can convert our position vector into a decimal vector that can be utilized further.
% \\For example, if the position vector of an individual generated by PSO is \(X_{id} = \{2.34, 4.63, -8.45, 0.64, -2.54\}\). This is a 5 dimension position vector. If the original order of fragments is \(S_{id} = [f_1, f_2, f_3, f_4, f_5]\), i.e. the Sequence Vector, then using SPV rule, we assign the first fragment to the shortest value in \(X_{id}\) and so on. Thus, the Sequence Vector for this particle becomes \(S_{id} = [f_3, f_5, f_4, f_2, f_1]\).
Figure \ref{SPV} shows an example of the SPV rule in the case of Genome Sequencing with a 5-dimensional particle.
\begin{figure}[h]
\centering
    \includegraphics[width=0.5\linewidth]{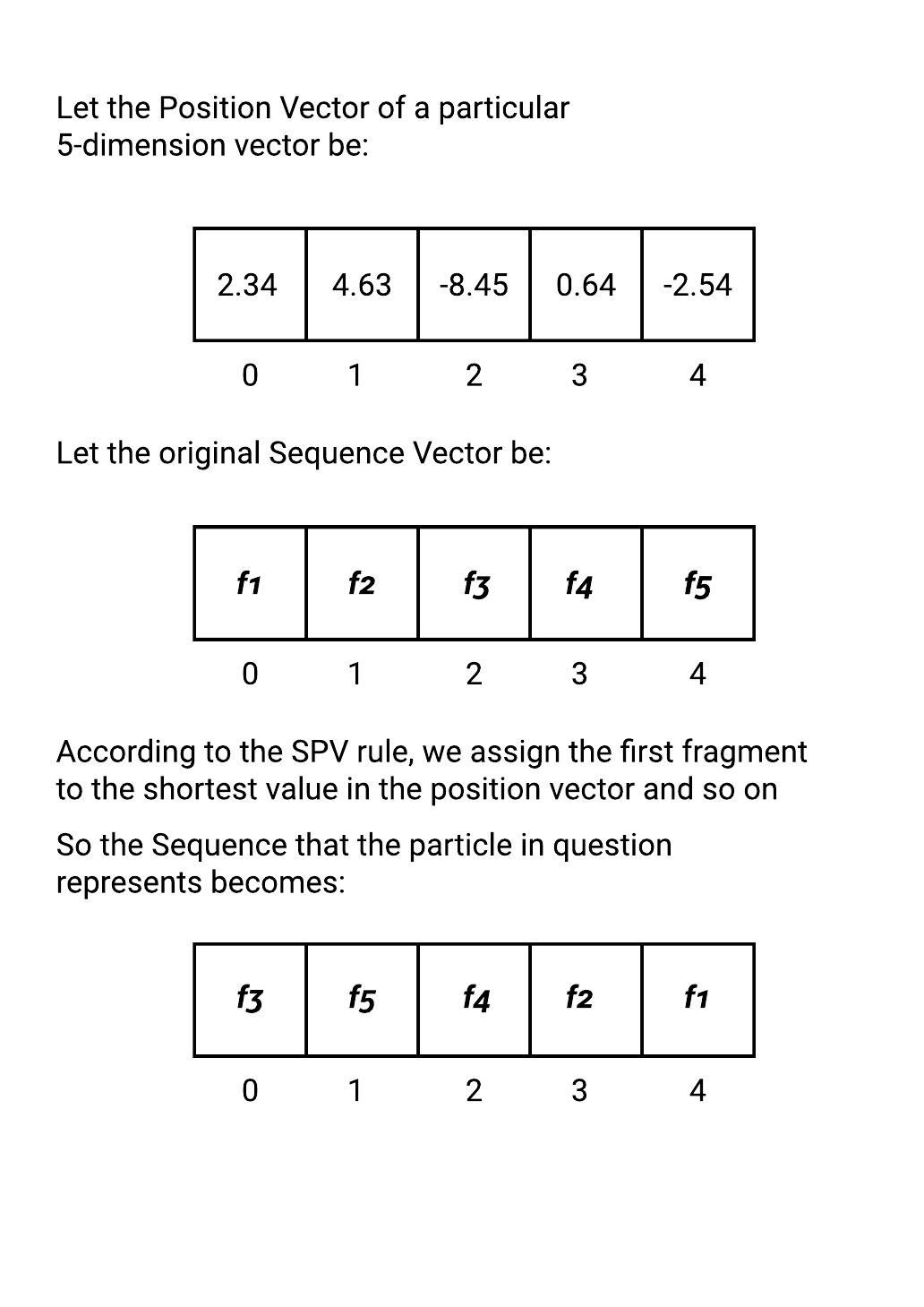}
    \caption{SPV rule in Genome Sequencing}
    \label{fig:SPV}
\end{figure}

\subsection{Fitness Function}
\label{Fitness}
Calculating the fitness value of a particle is crucial in determining which particle is doing well and ultimately drive the algorithm forward.
\\In the case of DNA Sequencing, one possible fitness function\cite{ShankarVerma2011}\cite{Parsons1995} can be the longest match between the suffix of one fragment and the prefix of the other. If \(n\) represents the length of the match then:
\begin{equation}
score_{i, i+1} = \begin{cases}
0 &\mbox{if } n \equiv 0 \\ 
n &\mbox{if } n > 0 \end{cases}
\end{equation}

Here, \(score_{i, i+1}\) represents the score of 2 paired fragments. The sum of the fragment pair score is calculated to compute the fitness of a particular particle generated in the PSO algorithm.
\begin{equation}
fitness_i = \sum_{j=0}^{D-1} score_{j, j+1}
\end{equation}

This fitness represents the performance value of the particle. In the case of Genome Sequencing, this fitness value has to be maximized.

\subsection{Proposed Algorithm}
The proposed variant of PSO consists of three parts, Chaos, time-varying coefficients and Levy Flight. It makes use of the SPV rule to convert a permutation based problem into a discreet problem.
\subsubsection{Chaos}
Chaos has been incorporated in two ways in the proposed algorithm, using Chaotic Inertia Weight and using Chaotic initialization.

As discussed in Section \ref{chaos}, a chaotic random Inertia Weight approach has been considered.
A lower Inertia Weight encourages exploitation, and a higher Inertia Weight favours exploration. A chaotic Inertia Weight ensures a balance between both these aspects. Various Chaotic Maps can be used here.

The logistic map\cite{May1976},
\(z_{t+1} = \mu z_t(1-z_t) \), is a very common chaotic map. However, it fails to guarantee chaos in the initial values of \(z_0\).
In this paper, the Sine Chaotic Map\cite{Hua2019} has been utilized\cite{Hong2016,Feng2007}.
\begin{equation}
    z_{t+1} = \beta sin(\pi z_t)
\end{equation}
where \(\beta>0, z_t, z_{t+1} \epsilon [0,1]\) and t is the generation number.

\begin{figure}[h]
    \centering
    \includegraphics[]{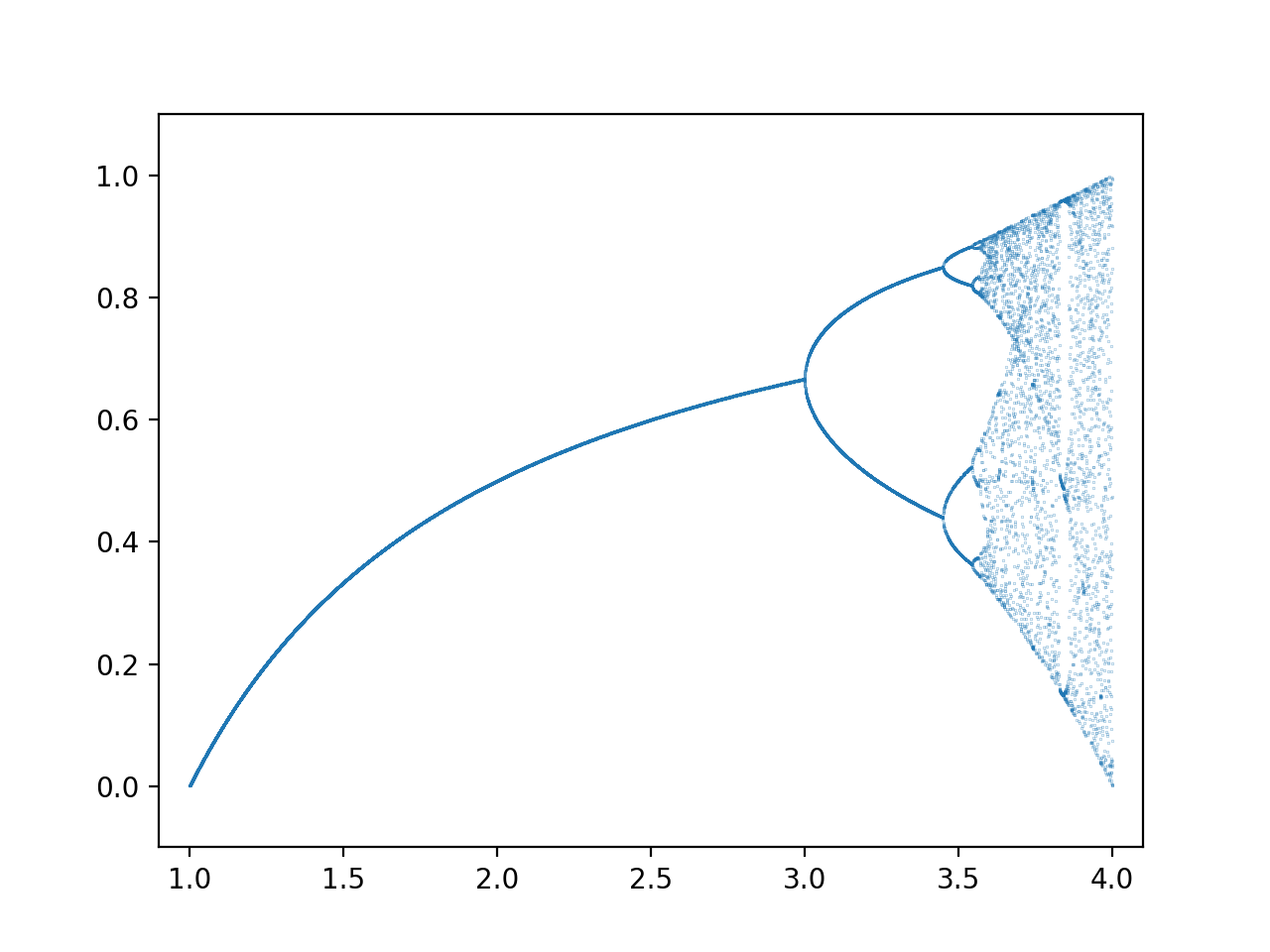}
    \caption{Bifurcation graph of Sine Chaotic Map}
    \label{fig:sineMap}
\end{figure}

Figure \ref{fig:sineMap} shows the bifurcation graph of the Chaotic Sine Map.

The following modification is done to improve the effectiveness of this map\cite{Hong2016}.

\begin{equation}
    z_{t+1} = \abs[\Bigg]{sin(\frac{\pi z_t}{rand(.)})}
\end{equation}
where \(\beta = 1, , z_t, z_{t+1} \epsilon [0,1]\). 
Thus, the final equation for inertia weight becomes:
\begin{equation}
    \omega_{chaos}^t = 0.5 * rand(.) + 0.5 * z_{t+1}
\end{equation}

Chaos has also been incorporated in the initialization of the particles for the proposed algorithm. Generation of particles with Chaos and refinement using Levy Flight (Refer Section \ref{levy2}) ensures the initialization of the particles with a high fitness score. 
Since by incorporating Chaotic Inertia, the swarm is highly influenced by its starting parameters and particles, this chaotic initialization ensures that the starting conditions favour finding better values.

\subsubsection{Time varying Coefficients}
A Constriction Coefficient\cite{Shi2000} has also integrated to prevent the divergence of the particles during search for solution as follows:
\begin{equation}
    \chi = \frac{2}{\phi -2 + \sqrt{\phi^2 -4\phi}}
\end{equation}

where \(\phi = c_1 +c_2\).

We have also incorporated time varying social and cognitive coefficients for improving local and global search\cite{Hong2016}.
A decreasing Cognitive Co-efficient and an increasing Social Coefficient enhance the exploration and exploitation of the PSO, thus getting the best convergence results. The mathematical formulation of the equation is given by:
\begin{equation}
    c_1^t = c_{1,f} - \frac{t}{MAXITR}\abs{(c_{1,f} - c_{1,i)}{}}
\end{equation}
\begin{equation}
        c_2^t = c_{2,i} + \frac{t}{MAXITR}\abs{(c_{2,f} - c_{2,i)}{}}
\end{equation}

where \(c_{1,i}\), \(c_{1, f}\), \(c_{1, i}\) and \(c_{2, f}\) are the initial and final values of the cognitive and social coefficients respectively, \(t\) is the current generation number and MAXITR is the number of iterations that the algorithm is being run for.
\subsubsection{Levy Flight}
\label{levy2}
Levy Flight has been used here to counter stagnancy, a condition that occurs when a particle gets stuck in a particular region without any improvement to its personal best for a long period of time. A particle if found to be in such a condition, undergoes Levy Flight random walk to find its new position.
Breaking down Levy Flight into its bare essentials,
\begin{enumerate}
    \item There’s a high probability that after doing Levy Flight, the object in question would be somewhere close to its original position.
    \item There’s a very low probability that after doing Levy Flight, the object would cover a far greater distance and abandon the neighborhood it was in.
\end{enumerate}

This ensures that the algorithm and its particles do not get stuck at any point. Once a particle remains stagnant for a set number of iterations, it is made to do Levy Flight to ensure that it gets out of the local optimum.
\subsubsection{Confluence}
\begin{figure*}[p]
    \centering
    \includegraphics[width=\textwidth,height=\textheight - 26.85pt,keepaspectratio,trim=4 4 4 4,clip]{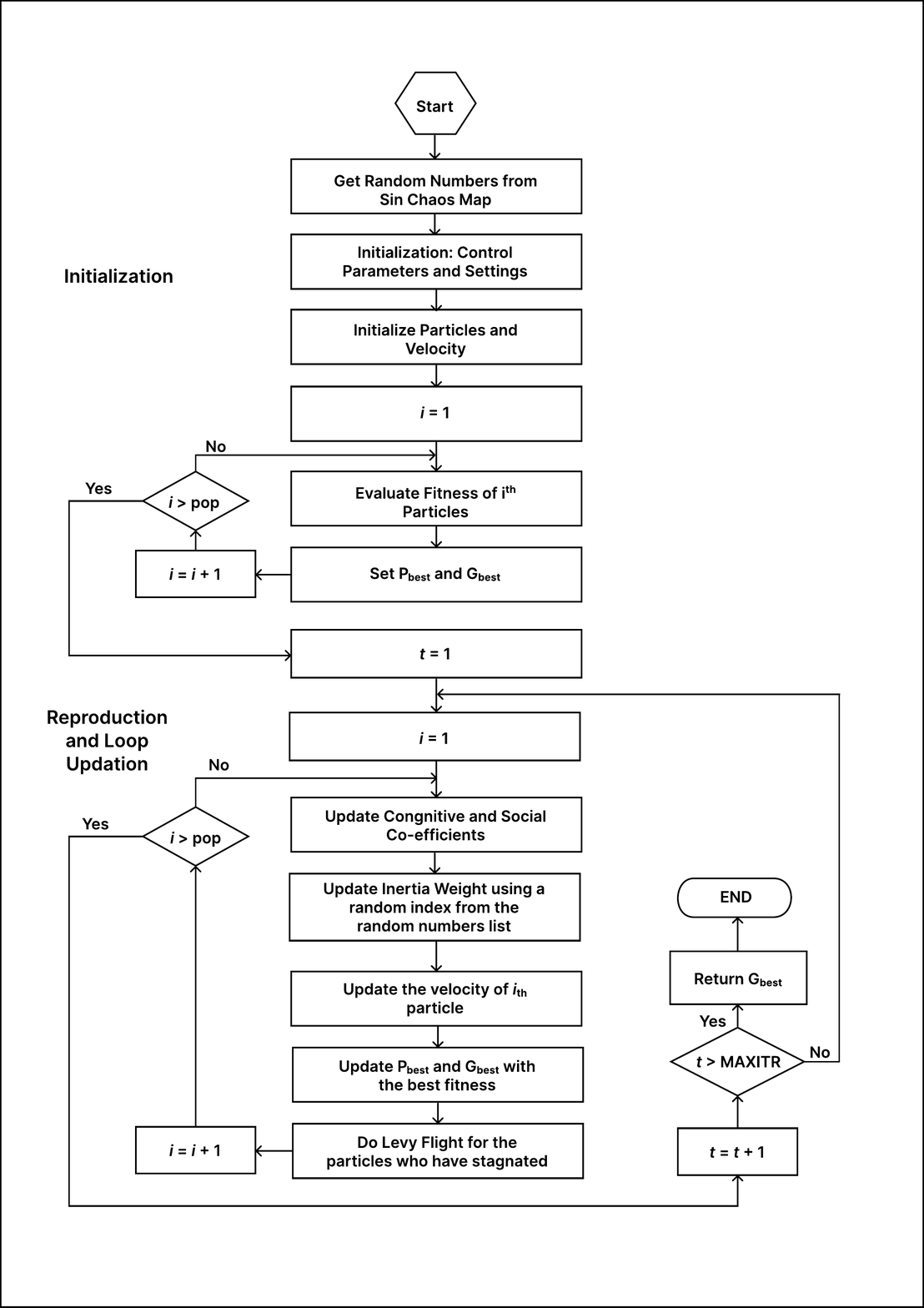}
    \caption{Flowchart for the proposed Algorithm}
    \label{fig:flowchart}
\end{figure*}

Bringing both these ideas together, the final equation of the PSO algorithm become:
\begin{equation}
    v_p^{t+1} = \chi * \omega_{chaos}^t * v_p^t + c_1^t * (p_{best}^t -x_{p}^t) + c_2^t * (g_{best}^t -x_{p}^t)
\end{equation}
\begin{equation}
    x_p^{t+1} = x_p^t + v_p^{t+1}
\end{equation}
The random numbers \(r_1\)and \(r_2\) from the velocity equation in the standard PSO have not been included.
Figure \ref{fig:flowchart} represents the flowchart and Algorithm \ref{algo:CPSOLF} represents the pseudo-code of the proposed Chaotic inspired PSO with Levy Flight.
\begin{algorithm}
\label{algo:CPSOLF}
\caption{Chaotic Particle Swarm Optimization with Levy Flight}
\begin{algorithmic} 
\STATE $max\_itr \leftarrow $\textit{Number of iterations}
\STATE $population \leftarrow $\textit{Swarm size}
\STATE $max\_stagnancy \leftarrow $\textit{Max Stagnancy}
\FOR{$i \leftarrow (1$\textit{ to population)}}
\STATE $x_i \leftarrow $\textit{initialize the position of} $x^{th}$ \textit{particle} 
\STATE $p_{best} \leftarrow x_i$
\STATE $Stagnancy_i \leftarrow 0$
\IF{$p_{best} > g_{best} $}
\STATE $G_{best} \leftarrow x_i$
\ENDIF
\ENDFOR
\STATE $t \leftarrow 0$
\WHILE{$t < max\_itr$}
\FOR{$i \leftarrow 1$ \textit{to population}}
\STATE $w \leftarrow$\textit{Chaotic Random Inertia Weight} 
\STATE $c_1^t \leftarrow c_{1,f} - \frac{t}{MAXITR}\abs{(c_{1,f} - c_{1,i)}{}}$
\STATE $c_2^t \leftarrow c_{2,i} + \frac{t}{MAXITR}\abs{(c_{2,f} - c_{2,i)}{}}$
\STATE $v_{id} \leftarrow w * v_{id} + c_1^t(p_{id}- x_{id}) + c_2^t(p_{gd} -  x_{gd})$
\STATE $x_i(t) \leftarrow x_i(t-1) + v_i$
\IF{$fitness(x_i) > fitness(p_{best}i)$}
\STATE $p_{best}i \leftarrow x_i$
\IF{$fitness(p_{best}i) > fitness(g_{best})$}
\STATE $Stagnancy_i \leftarrow 0$
\STATE $g_{best} \leftarrow p_{best}i$
\STATE $g_{best} \leftarrow x_i$
\ENDIF
\ELSE 
\STATE $Stagnancy_i \leftarrow Stagnancy_i + 1$
\ENDIF
\FOR{$i \leftarrow  1$ \textit{to population}}
\IF{$Stagnancy_i < max\_stagnancy$}
\STATE $LevyFlight(p)$
\ENDIF
\ENDFOR
\ENDFOR
\STATE $t \leftarrow t + 1$
\ENDWHILE
\end{algorithmic}
\end{algorithm}
\section{Experiment Results}
\label{Exp Results}
In this section, the performance comparison of the proposed method in comparison to the state-of-arts algorithms is presented. The NCBI\footnote{http://www.ncbi.nlm.nih.gov} Sequence Read Archive\footnote{https://www.ncbi.nlm.nih.gov/sra} are used for the analysis. All algorithms were implemented using Python 3.7.9 and performed on a 3.1 GHz Dual-Core Intel Core i5 with 8GB of memory running Mac OS 11.2.

\subsection{Datasets}
The following four datasets were taken from the NCBI Short Reads Archive:1. \begin{enumerate}
    \item SRR13493107 : 16S rRNA of Homo sapiens:  human fecal microbiome
    \item SRR13660057 : PCR Tiled Amplification of SARS-CoV-2 1 ILLUMINA (Illumina MiSeq) run: 27,125 spots, 7.8M bases
    \item SRR13660059 : PCR  Tiled  Amplification  of  SARS-CoV-2  1 ILLUMINA (Illumina MiSeq) run:  28,228 spots, 8.1M bases
    \item SRR13660101  :   PCR  Tiled  Amplification  of  SARS-CoV-2  1 ILLUMINA (Illumina MiSeq) run:  27,949 spots, 8.1M bases
\end{enumerate}

\subsection{Parameters}
\label{parameters}
Six different variants of Particle Swarm Optimization were implemented and then compared with the proposed algorithm. The list of algorithms is as follows:
\begin{enumerate}
    \item Chaotic PSO with Levy Flight (CPSOLF)
    \item Simple PSO (SPSO)
    \item SPSO with Inertia (SPSOI)
    \item SPSO with dynamic Inertia (SPSODI)
    \item Accelerated PSO (APSO)
    \item APSO with Inertia (APSOI)
    \item APSO with dynamic Inertia (APSODI)
\end{enumerate}
For all the PSO algorithms, parameter were fixed for experimental analysis. For an unbiased analysis of the performance of all algorithms, the number of iterations and the population size were fixed. Table \ref{tab:parameters} represents the parameters that have been experimentally set up.

\begin{table}
\centering
\caption{Parameters}
\label{tab:parameters}
\begin{tabular}{ll}
\hline\noalign{\smallskip}
Parameter & Value   \\
\noalign{\smallskip}\hline\noalign{\smallskip}
Number of Iterations & 50\\
Population & 10\\
Range if Linearly Decreasing Inertia Weight & 0.9 - 0.4\\
Constant Inertia Weight & 0.7\\
Linearly Varying Social Co-efficient & 1.7 - 2.4\\
Linearly Varying Cognitive Co-efficient & 2.4 - 1.7\\
Constant Cognitive Co-efficient & 2\\
Constant Social Co-efficient & 2\\

\noalign{\smallskip}\hline
\end{tabular}
\end{table}

\subsection{Results}
Table \ref{tab:result} represents the outputs of the different variants of PSO. These tables include Mean, Standard Deviation, Best and Worst values of \(g_{best}\) when run 10 times on each dataset.

It is evident from Table \ref{tab:result} that the proposed algorithm CPSOLF, offers a consistently better mean score among all the variations of Particle Swarm Optimization considered.
A possible reason of this improvement may be the good exploration and exploitation of the given search space using chaos and levy flight. 
\begin{table*}[htb]
\small
\centering
    \caption{Results of Algorithms}
    \begin{tabular}{ccccccccc}
    \noalign{\smallskip}\hline
Datasets & \multicolumn{8}{c}{Algorithms} \\
& & \textbf{CPSOLF} & SPSO & SPSOI & SPSODI & APSO & APSOI & APSODI \\
\noalign{\smallskip}\hline
SRR13660101 & Mean & \textbf{113,303.00} & 105,516.40 & 106,316.00 & 111,522.40 & 102,145.00 & 99,244.40 & 99,047.90 \\
& SD & \textbf{15,410.14} & 13,613.74 & 17,603.19 & 22,287.84 & 11,713.31 & 10,611.53 & 11,836.93 \\
& Best & \textbf{131,343.00} & 131,058.00 & 132,811.00 & 157,316.00 & 124,637.00 & 117,648.00 & 115,592.00 \\
& Worst & \textbf{84,812.00} & 90,766.00 & 89,571.00 & 91,823.00 & 83,866.00 & 83,144.00 & 81,999.00 \\\noalign{\smallskip}\hline
SRR13493107 & Mean & \textbf{38,530.60} & 38,418.90 & 38,470.60 & 38,454.50 & 38,487.70 & 38,459.90 & 38,434.10 \\
& SD & \textbf{120.88} & 64.76 & 83.41 & 52.55 & 117.35 & 93.52 & 73.00 \\
& Best & \textbf{38,769.00} & 38,491.00 & 38,603.00 & 38,526.00 & 38,661.00 & 38,649.00 & 38,532.00 \\
& Worst & \textbf{38,360.00} & 38,327.00 & 38,348.00 & 38,374.00 & 38,338.00 & 38,346.00 & 38,301.00 \\\noalign{\smallskip}\hline
SRR13660057 & Mean & \textbf{161,271.30} & 156,059.10 & 132,180.60 & 142,026.80 & 156,966.40 & 158,391.90 & 131,112.30 \\
& SD & \textbf{9,645.95} & 21,011.64 & 35,808.81 & 30,532.12 & 14,696.65 & 21,011.64 & 34,397.43 \\
& Best & \textbf{184,570.00} & 172,001.00 & 171,250.00 & 169,030.00 & 191,972.00 & 200,724.00 & 162,591.00 \\
& Worst & \textbf{151,286.00} & 108,540.00 & 86,312.00 & 85,766.00 & 140,914.00 & 112,948.00 & 89,095.00 \\\noalign{\smallskip}\hline
SRR13660059 & Mean & \textbf{216,958.10} & 187,009.40 & 193,378.80 & 205,007.20 & 203,122.50 & 182,311.20 & 202,957.70 \\
& SD & \textbf{32,005.91} & 28,041.58 & 28,855.44 & 28,031.27 & 28,542.32 & 16,578.68 & 33,956.77 \\
& Best & \textbf{267,298.00} & 258,277.00 & 242,439.00 & 242,908.00 & 253,842.00 & 222,504.00 & 282,363.00 \\
& Worst & \textbf{159,033.00} & 164,147.00 & 164,297.00 & 166,666.00 & 164,298.00 & 167,550.00 & 166,772.00 \\\noalign{\smallskip}\hline
\end{tabular}

    \label{tab:result}
\end{table*}
\begin{figure}[h]
    \centering
    \includegraphics[width=\linewidth-40pt,height=\textheight - 26.85pt,keepaspectratio]{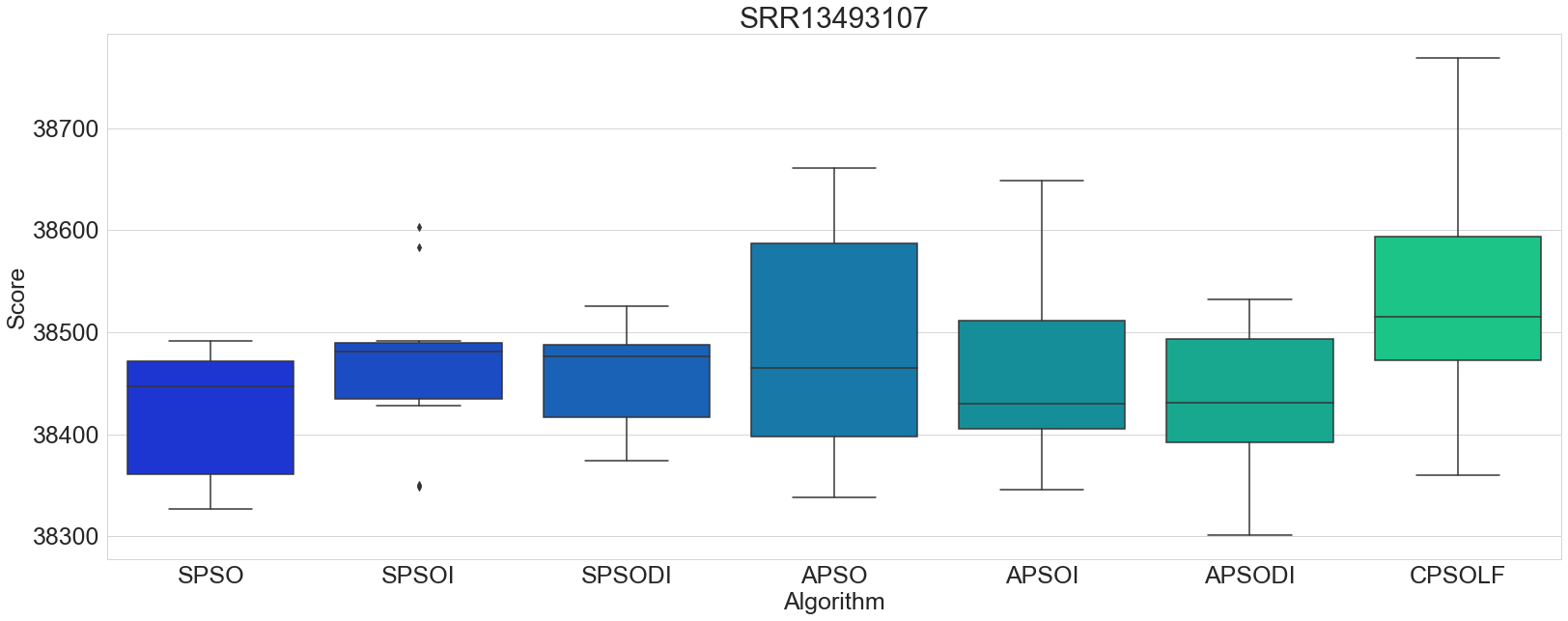}
    \caption{SRR13493107 Comparison}
    \label{fig:SRR13493107}
\end{figure}

\begin{figure}[H]
    \centering
    \includegraphics[width=\linewidth-40pt,height=\textheight - 26.85pt,keepaspectratio]{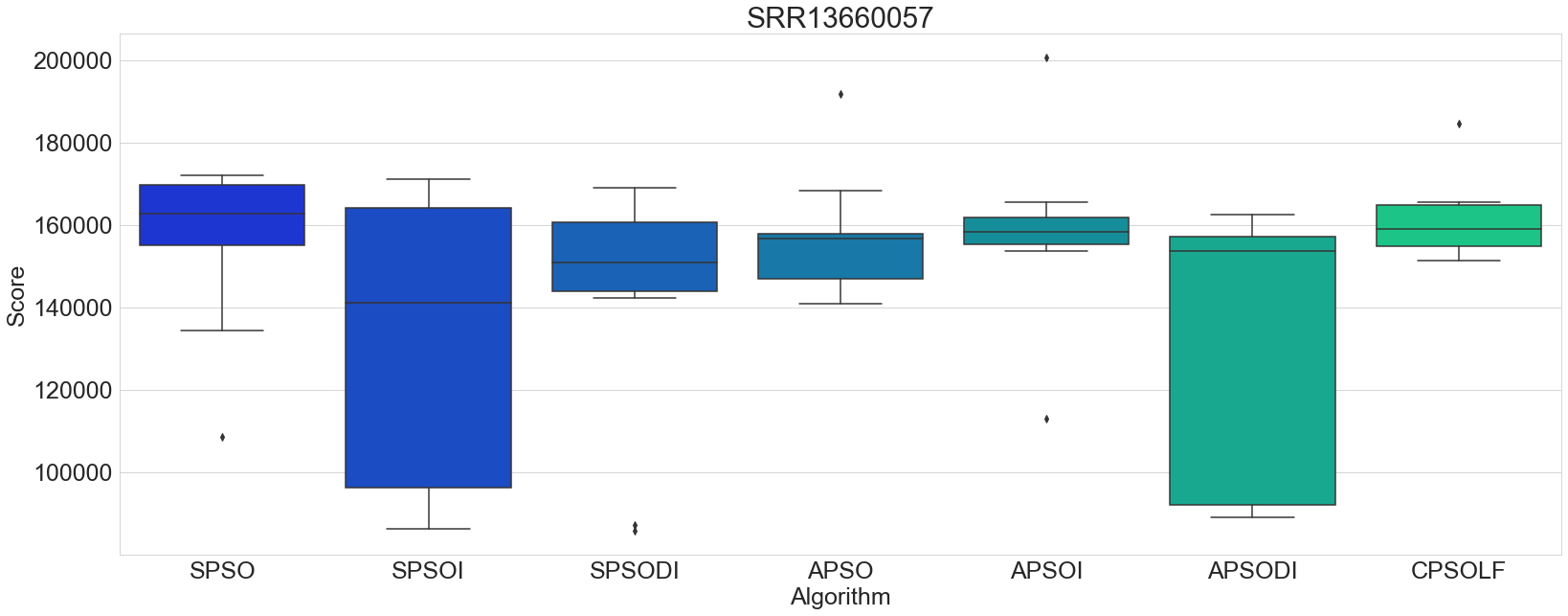}
    \caption{SRR13660057 Comparison}
    \label{fig:SRR13660057}
\end{figure}

\begin{figure}[H]
    \centering
    \includegraphics[width=\linewidth-40pt,height=\textheight - 26.85pt,keepaspectratio]{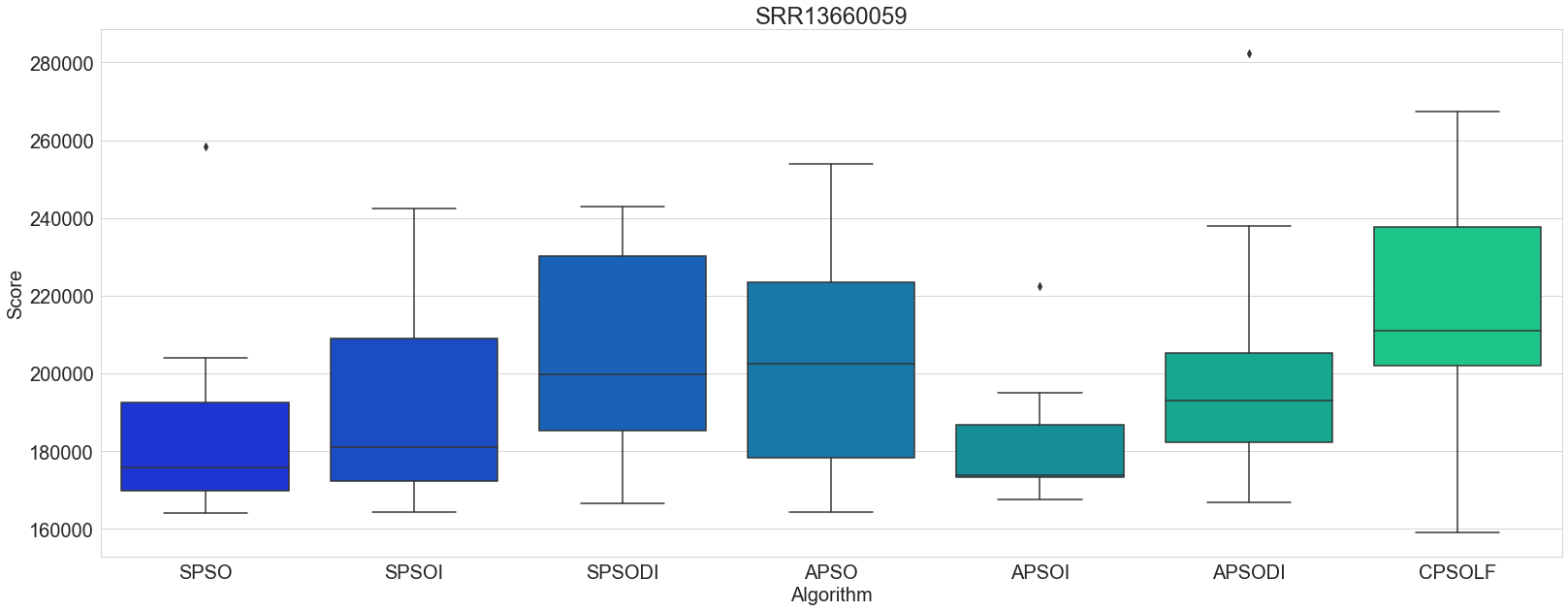}
    \caption{SRR13660059 Comparison}
    \label{fig:SRR13660059}
\end{figure}

\begin{figure}[H]
    \centering
    \includegraphics[width=\linewidth-40pt,height=\textheight - 26.85pt,keepaspectratio]{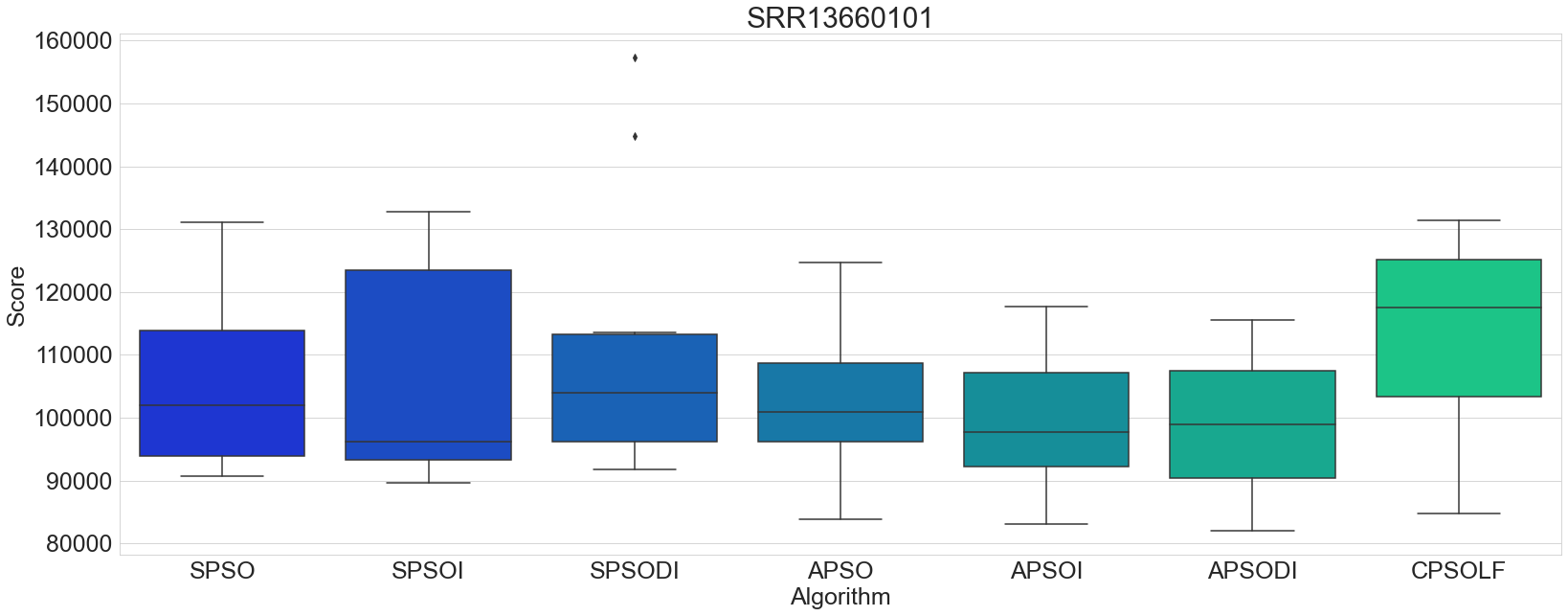}
    \caption{SRR13660101 Comparison}
    \label{fig:SRR13660101}
\end{figure}
Figures \ref{fig:SRR13493107}, \ref{fig:SRR13660057}, \ref{fig:SRR13660059}  and \ref{fig:SRR13660101} represent the box-plots for the algorithms. These plot have been included to demonstrate the comparative analysis of all models over different data sets.

\begin{figure}[H]
    \centering
    \includegraphics[width=\linewidth-20pt,height=\textheight - 26.85pt,keepaspectratio]{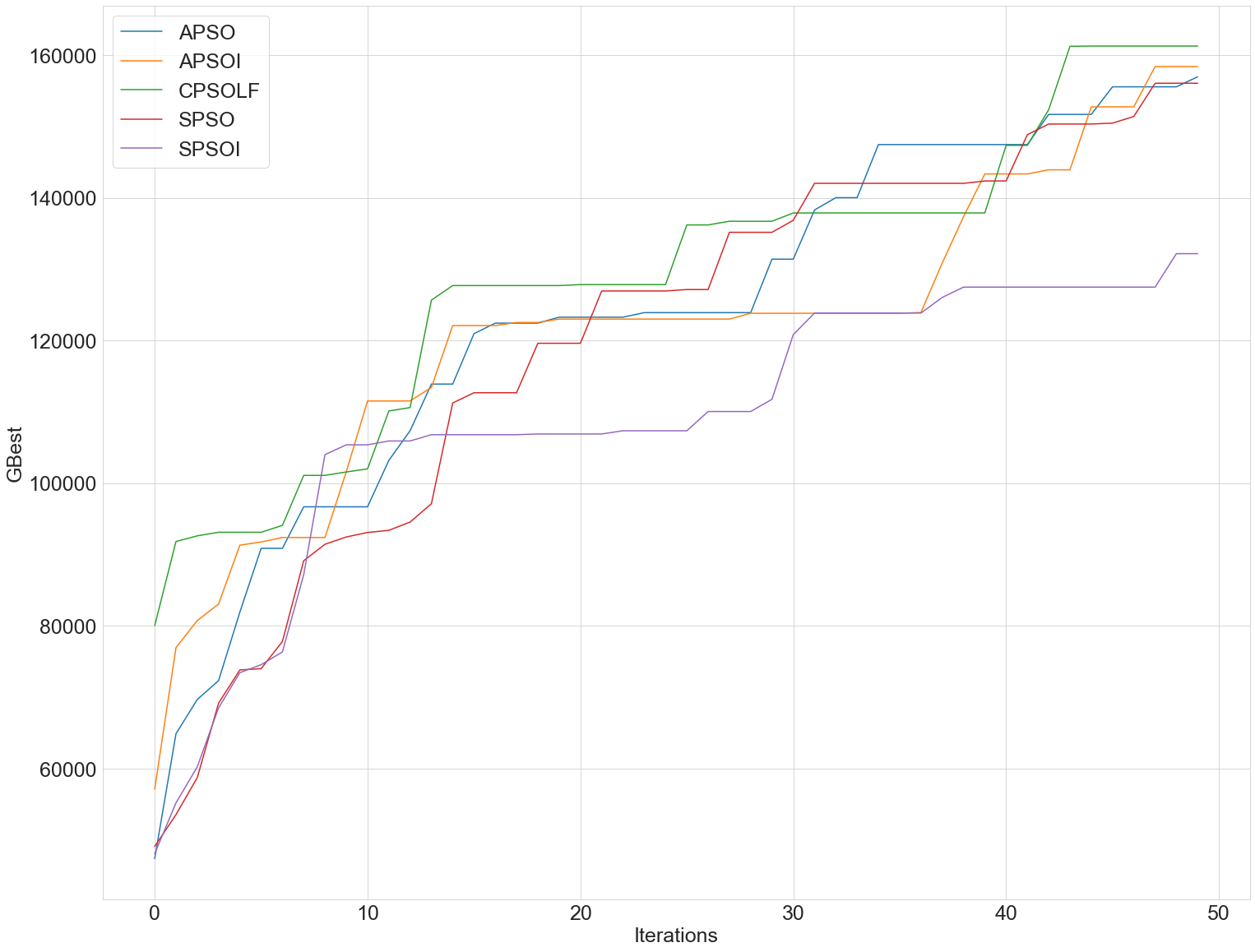}
    \caption{Convergence of the algorithms}
    \label{fig:Convergence}
\end{figure}

 Figure \ref{fig:Convergence} shows the convergence of the various algorithms.

% {\color{red} Make your convergence line more wide. Make alignment between text , figure and table. They should come one after the other. Information about table 1 should be around table one only}
Figure \ref{fig:Convergence} shows that the proposed algorithm, CPSOLF (Green), outperforms the other variants over the iterations.
The use of Chaos ensured that it commenced with more favourable conditions and converged at a higher fitness than other versions.

Exploration and Exploitation are two facets of PSO's local search. The downside of the PSO algorithm is that it is easy to fall into a local maximum in a high-dimensional space and has a poor iterative convergence rate.
We substitute some stagnant particles with new particles by integrating Levy Flight into PSO. PSO has a better convergence rate this way.
\\In this case, chaos is also beneficial. The algorithm escapes stagnation and premature convergence by using Chaotic Inertia Weights and can escape from a local minimum. As a result, our proposed model outperforms all other competitive methods in all ways.

\subsection{Statistical Significance}

It is important to compare the statistical significance of the data collected using the proposed algorithm to the algorithms considered in its competition. Z-Tests have been used to determine this importance. Where the variances are known and the sample size is high, this statistical test is used to assess if two population means differ. In order to do an effective z-test, the test statistic is supposed to have a normal distribution, and nuisance parameters such as standard deviation should be known.\\These scores are listed in Table \ref{tab:stat}. These scores have been calculated considering a significance level ($\alpha = 0.05$). P Values have been calculated for each pair among the considered variants of PSO. P-values less than \(\alpha\) represent that the difference between means of the two algorithms is statistically significant.
As shown in table \ref{tab:stat}, CPSOLF shows a significant difference in 19 out of 24 experiments, performing significantly better in 3 of the 4 datasets considered.
\begin{table}[H]
\small
\centering
   \caption{Comparison of CPSOLF using Z-Tests on the Four Datasets}
    \begin{tabular}{ccccc}
    \noalign{\smallskip}\hline
Pairs & I & II & III & IV \\
\noalign{\smallskip}\hline
CPSOLF-SPSO & 0.00 & 0.00 & 0.04 & 0.01 \\
CPSOLF-SPSOI & 0.01 & 0.06 & 0.00 & 0.34 \\
CPSOLF-SPSODI & 0.12 & 0.02 & 0.00 & 0.02\\
CPSOLF-APSO & 0.09 & 0.13 & 0.08 & 0.04 \\
CPSOLF-APSOI & 0.00 & 0.03 & 0.17 & 0.09 \\
CPSOLF-APSODI & 0.08 & 0.01 & 0.00 & 0.03\\
\noalign{\smallskip}\hline
\end{tabular}
    \label{tab:stat}
\end{table}

\section{Conclusion and Future Work}
\label{conclusion}

Genome Sequencing is a very major field of research for modern science. Genome Sequencing, and its current algorithms, is highly constrained by availability of computing power and the huge time taken to sequence a genome of an organism.
This paper proposes a novel technique which has shown promising results for the Genome Sequencing Problem. By using Chaos, Levy Flight and Adaptive Parameters in PSO, it has a higher chance of succeeding where PSO cannot produce good results. The permutation-optimization problem is converted to a discrete optimization problem using the SPV rule.

The Chaotic Particle Swarm Optimization with Levy Flight outperforms other variants of the PSO. It shows significant difference from them which was calculated using Z-Tests. It performs 7\% to 24\% better when compared to other variants for the four datasets in consideration. It can be inferred that the proposed variant produces more optimum results when compared to other variants of Particle Swarm Optimization and is more reliable.

The proposed algorithm does not show much improvement when compared with others on the basis of Standard Deviation. It ranks higher however it lacks consistency. Other variants of the PSO algorithm can be implement to tackle this problem.

%\begin{acknowledgements}
%\end{acknowledgements}

\section*{Conflict of interest}

The authors declare that they have no conflict of interest.

 \bibliographystyle{elsarticle-num} 
\bibliography{GenomeSequencing.bib}   % name your BibTeX data base

\end{document}